\documentclass{article}
\usepackage{abstract}
\usepackage{amsmath,amssymb}
\usepackage{graphicx}
\usepackage{hyperref}
\usepackage{geometry}
\title{\textbf{KV-Efficient VLA: A Method to Speed up Vision Language Models with RNN-Gated Chunked KV Cache}}
\author{
  Wanshun Xu, Long Zhuang\\
  University of Toronto\\[1em]
  Lianlei Shan\\
  Tsinghua University\\[1em]
  \textbf{Github:} \href{https://github.com/l-zhuang/KV-Efficient-VLA}{KV-Efficient-VLA.github.io}
}
\date{}

\begin{document}

\maketitle
\renewcommand{\abstractname}{\normalfont\Large\bfseries Abstract}
\begin{abstract}
Vision--Language--Action (VLA) models offer a unified framework for robotic perception and control, but their ability to scale to real-world, long-horizon tasks is limited by the high computational cost of attention and the large memory required for storing key--value (KV) pairs during inference, particularly when retaining historical image tokens as context. Recent methods have focused on scaling backbone architectures to improve generalization, with less emphasis on addressing inference inefficiencies essential for real-time use. In this work, we present KV-Efficient VLA, a model-agnostic memory compression approach designed to address these limitations by introducing a lightweight mechanism to selectively retain high-utility context. Our method partitions the KV cache into fixed-size chunks and employs a recurrent gating module to summarize and filter the historical context according to learned utility scores. This design aims to preserve recent fine-grained detail while aggressively pruning stale, low-relevance memory. Based on experiments, our approach can yield an average of 24.6\% FLOPs savings, 1.34$\times$ inference speedup, and 1.87$\times$ reduction in KV memory. Our method integrates seamlessly into recent VLA stacks, enabling scalable inference without modifying downstream control logic.
\end{abstract}


\section{Introduction}

Recently, vision--language models (VLMs)~\cite{caron2021emergingpropertiesselfsupervisedvision,radford2021learningtransferablevisualmodels,zhai2023sigmoid,alayrac2022flamingo} have demonstrated strong capabilities in instruction following, visual grounding, and semantic reasoning by leveraging internet-scale image--text pretraining. Building on these successes, recent efforts have extended VLMs to vision--language-action (VLA) models~\cite{ahn2022can,kim2024openvla,li2024cogact,liu2025hybridvla,zitkovich2023rt}, allowing robotic agents to interpret visual and linguistic inputs and execute the corresponding actions in physical environments. These models typically integrate multimodal perception with policy learning within transformer-based architectures, allowing generalized, instruction-driven robotic behavior~\cite{embodimentcollaboration2025openxembodimentroboticlearning,khazatsky2025droidlargescaleinthewildrobot}. As robotic systems increasingly operate in diverse and unstructured environments, there is a growing need for VLA models that can operate efficiently in real time~\cite{james2019rlbenchrobotlearningbenchmark,wu2025robomindbenchmarkmultiembodimentintelligence}.

Earlier work on VLA demonstrated promising generalization, but they faced practical challenges related to computational efficiency. In particular, maintaining complete key--value (KV) caches across layers and time steps can lead to increased memory consumption and latency, especially as action sequences grow longer. In diffusion-based components, repeated sampling introduces multiple rounds of visual--language fusion, which can compound computational overhead. For instance, OpenVLA (7B)~\cite{kim2024openvla}, employing purely autoregressive decoding, operates at approximately 6~Hz. HybridVLA (7B), which combines both autoregressive decoding and diffusion-based planning, performs at 6.1~Hz~\cite{liu2025hybridvla} due to the added overhead of repeated sampling. These inference speeds fall short of the $>50$--$100$~Hz rates typically required for smooth and responsive control in real-time robotics tasks such as dexterous manipulation or dynamic navigation~\cite{li2024survey}. Past studies have observed that, in VLA systems, frames contain redundant information. Including raw historical frames without refinement has been shown to reduce performance by approximately 6\%, highlighting the need for more efficient temporal context integration~\cite{zheng2024tracevla}.

\begin{figure}[ht]
    \centering
    \includegraphics[width=0.65\textwidth]{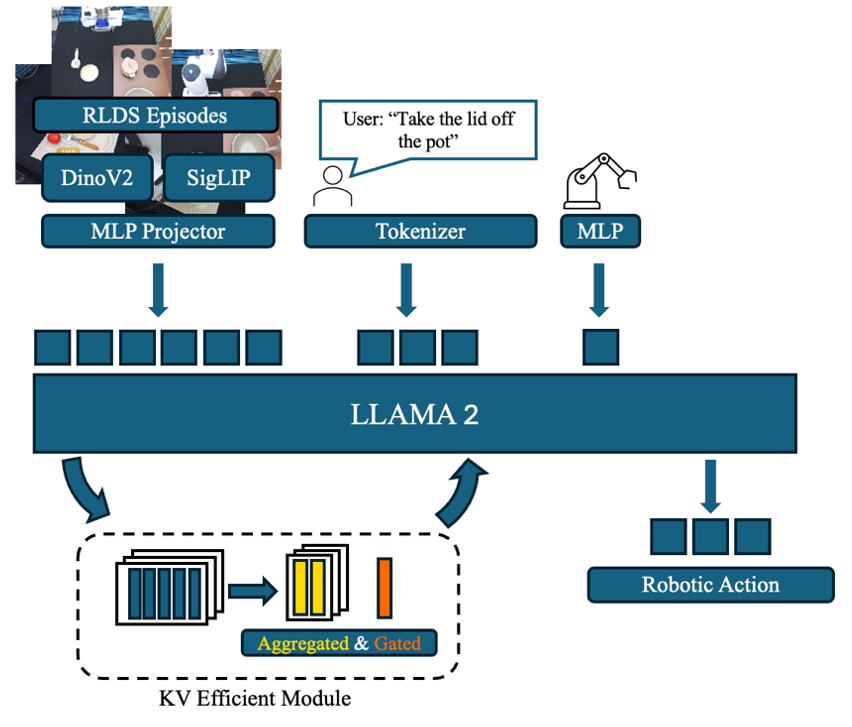}
    \caption{VLA is pretrained on diverse robotic datasets, incorporating the KV Efficient Module. The input data consists of prompts, images and robotic state, while the output comprises robotic action vectors.}
    \label{fig:overall}
\end{figure}

To alleviate these runtime and memory bottlenecks, we propose a lightweight KV-Efficient module (Figure~\ref{fig:overall}) designed to accelerate attention in long-horizon action generation by compressing and selectively retaining context. Specifically, the raw KV cache is divided into fixed-length chunks, with each chunk aggregated and processed through an LSTM module~\cite{LSTM}. We employ the pretrained LLAMA 2 7B~\cite{LLAMA2}, replacing the traditional KV cache with the KV-Efficient module. Fine-tuning is conducted using low-rank adaptation (LoRA)~\cite{Lora} to compensate for approximation error, implemented through the LLAMA Factory framework~\cite{zheng2024llamafactory} with robotic datasets, Open X-Embodiment~\cite{embodimentcollaboration2025openxembodimentroboticlearning}. By applying our KV-efficient attention design to existing VLA architectures, we achieve up to 1.34$\times$ faster inference speed on average compared to baseline models. Our contributions include:

\begin{itemize}
    \item \textbf{Chunked KV Strategy:} Reduces full-sequence caches by segmenting past tokens into fixed-size blocks and aggregating with an MLP.
    \item \textbf{LSTM Gating Mechanism:} Retains only the most informative aggregated blocks using an LSTM gate.
\end{itemize}

\section{Related Works}

\textbf{Vision-Language-Action Models:}
VLA models unify vision, language, and action modalities into end-to-end systems for robotic control. Early methods such as SayCan~\cite{ahn2022can} and RT-2~\cite{zitkovich2023rt} employed pretrained vision--language models with planners to translate language instructions into grounded actions. More recent approaches like RT-2-X~\cite{embodimentcollaboration2025openxembodimentroboticlearning} and OpenVLA~\cite{kim2024openvla} fine-tune large-scale vision--language transformers to directly predict tokenized actions, enabling semantic-level planning and generalization. Diffusion-based models CogACT~\cite{li2024cogact} generate actions through iterative denoising, capturing rich multimodal distributions and enabling fine-grained control. HybridVLA~\cite{liu2025hybridvla} integrates an autoregressive planner for high-level intent and a diffusion-based decoder for action refinement, leveraging the strengths of both paradigms to improve semantic reasoning and execution fidelity.

\medskip
\noindent
\textbf{Vision-Language Models:}
VLMs based on transformer architectures have become fundamental to multimodal representation learning by aligning image and text features in a shared embedding space. CLIP~\cite{radford2021learningtransferablevisualmodels} pioneered large-scale contrastive pretraining, enabling strong zero-shot transfer across diverse visual tasks. SigLIP~\cite{zhai2023sigmoid} improves upon CLIP by replacing the SoftMax contrastive loss with a sigmoid-based objective, offering better training stability and retrieval performance, especially in smaller batch regimes. Self-supervised vision encoders like DINOv2~\cite{oquab2023dinov2} further enhance visual feature quality without labeled data, advancing state-of-the-art results in classification and segmentation. Models like Flamingo~\cite{alayrac2022flamingo} extend VLM capabilities through cross-attention-based fusion of visual and language tokens, enabling few-shot learning in multimodal settings.

\textbf{Transformer Attention:} The transformer autoregressively predicts the next action token
given a history of multimodal tokens (e.g., images, instruction)~\cite{vaswani2017attention}. At decoding step $t$,
each attention layer forms a query $Q_t \in \mathbb{R}^{H \times d_k}$ for the new token and attends
to stored keys/values $\{K_{1:t-3}, V_{1:t-3}\}$ from all prior tokens. With standard softmax
attention, the per-head attention over $n$ cached positions is

\begin{equation}
\label{eq:attention}
\text{Attention}(Q_t, K_{1:n}, V_{1:n}) =
\text{softmax}\left(\frac{Q_t K_{1:n}^\top}{\sqrt{d_k}}\right) V_{1:n}.
\end{equation}

Maintaining the full cache yields $\mathcal{O}(n)$ time and memory per new token per head, which
grows with context length and is the main bottleneck for real-time control.

\section{Methods}

Vision--Language--Action systems have advanced perception and policy learning, but inference-time scalability remains a bottleneck due to unbounded KV cache growth, which increases latency and memory usage in real-time control. To address this, we introduce the KV-Efficient module. Section~3.1 presents the overall framework and its role in scalable inference. Section~3.2 explains the KV cache chunking and recurrent gating strategy used to constrain cache growth. Section~3.3 analyzes the computational cost, quantifying the efficiency improvements of the proposed design.

\subsection{Overall Framework}

\begin{figure}[t]
    \centering
    \includegraphics[width=0.8\textwidth]{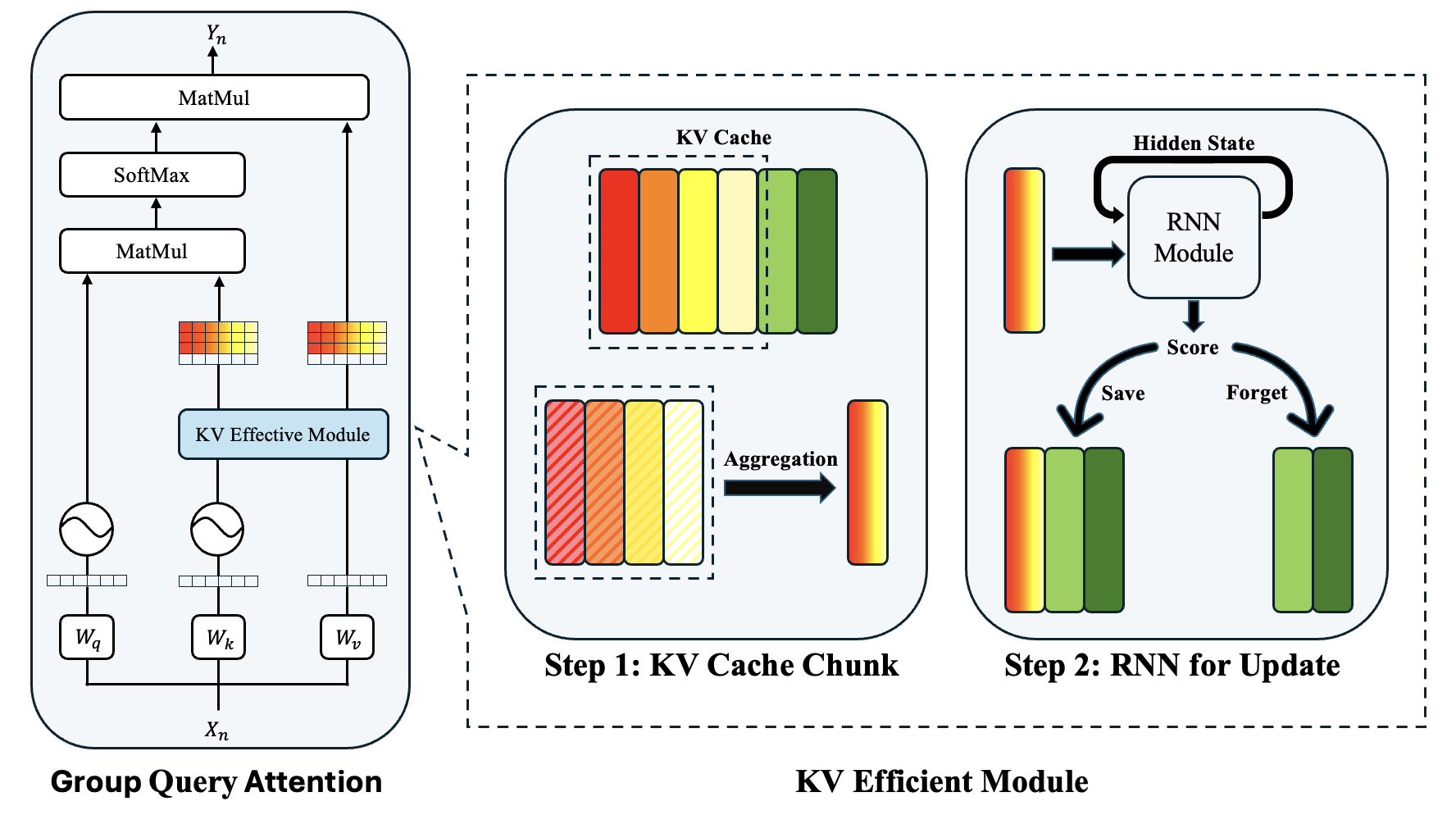}
    \caption{\textbf{KV-Efficient Framework.} The framework consists of two main steps: (1) Chunking the KV cache into fixed-size segments and aggregating them, and (2) Using an RNN module to update and select which chunks to retain or discard, enabling efficient and selective compression for attention mechanisms.}
    \label{fig:kvefficient}
\end{figure}

As illustrated in Figure~\ref{fig:kvefficient}, the proposed KV-Efficient framework first partitions the key--value cache generated by the attention mechanism into multiple chunks of fixed length. Each chunk is subsequently aggregated, for example through pooling, averaging, or a multi-layer perceptron~\cite{sitzmann2020implicit}, to produce a more compact representation. This aggregation step effectively reduces the memory footprint and prepares the data for selective retention. Next, each compressed chunk is passed through an LSTM, maintaining a hidden state across time or sequence steps. It evaluates each aggregated chunk and outputs a gating score that quantifies its importance. Based on this score, the framework dynamically decides whether to keep or forget each chunk for future processing. This selective update strategy strikes a balance between memory compression and information preservation, allowing the model to efficiently manage longer context windows during inference. Ultimately, the preserved aggregated chunks are attended to jointly, ensuring that essential information from the input sequence remains accessible to the model.

\subsection{KV Cache Chunking with Recurrent Gating}

The KV cache stores the key--value pairs $\{(K_t, V_t)\}$ in temporal order. To reduce memory usage, these pairs are partitioned into non-overlapping chunks of fixed length $C \in \mathbb{N}$. For each chunk at time $t$, defined as $C_t = \{(k_j,v_j)\}_{j=1}^{C}$, we compute a representative vector for both keys and values using a Multi-Layer Perceptron:

\begin{equation}
\label{eq:mlp_aggregate}
\bar{K}_t = \mathrm{MLP}_K(\{k_j\}_{j=1}^{C}),
\qquad
\bar{V}_t = \mathrm{MLP}_V(\{v_j\}_{j=1}^{C}),
\qquad
\bar{K}_t, \bar{V}_t \in \mathbb{R}^{B \times 1 \times d}
\end{equation}

After this aggregation step, each chunk is summarized into a single pair, $[\bar{K}_t, \bar{V}_t]$, with shape $B \times H \times 1 \times d$, where $B$ is the batch size and $d$ is the feature dimension. This process preserves the dimension of each attention head.

To decide which compressed chunks to keep, a recurrent gating mechanism is used. Specifically, the sequence of chunk representatives $([\bar{K}_t, \bar{V}_t])$ is passed through an LSTM gate with hidden size $d_g$, which maintains a hidden state $h_t$ and produces a gating score $s_t \in [0,1]$:

\begin{equation}
\label{eq:lstm_gate}
h_t,\, s_t = \mathrm{LSTM}([\bar{K}_t, \bar{V}_t],\, h_{t-1}).
\end{equation}

The cache update policy uses a learnable threshold $\tau$. If $s_t \geq \tau$, the compressed chunk will be retained in the cache; otherwise, it is discarded. To preserve fine-grained recency, the most recent $r < W$ tokens remain uncompressed. This update policy maintains strict causality, ensuring that at step $t$, the model attends only to past tokens $i < t$.

To clearly illustrate the proposed memory compression method, we provide the pseudocode as shown in Algorithm~1.

\begin{figure}[t]
\centering
\includegraphics[width=0.8\linewidth]{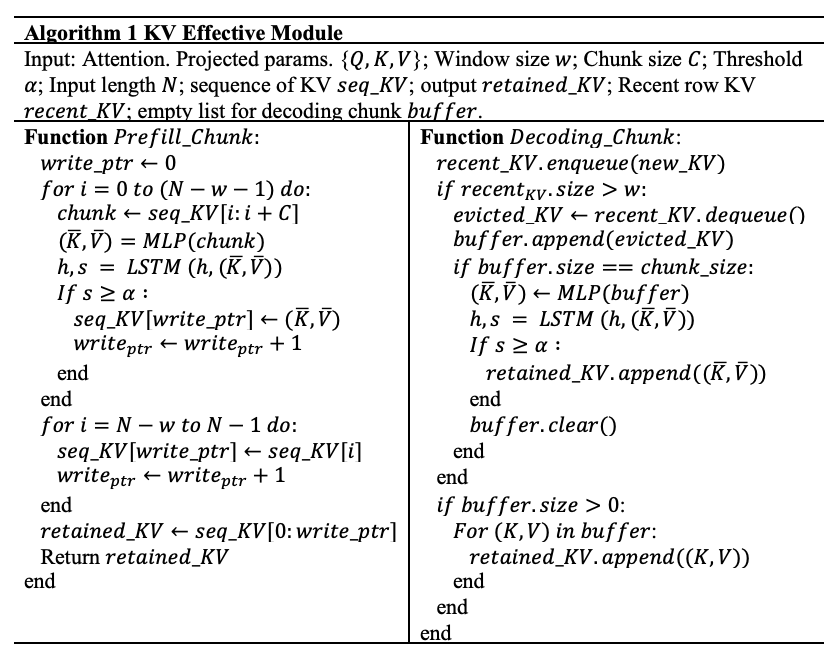}  
\end{figure}

\subsection{Computation Cost Analysis}

\textbf{Baseline:} For each new token, the layer performs linear projections to compute the query ($Q$), key ($K$), value ($V$), and output ($O$) matrices. Each of these operations involves a matrix multiplication of size $d_{\text{model}} \times d_{\text{model}}$, applied over a batch of size $B$, contributing a total computational cost of $4Bnd_{\text{model}}^{2}$. Attention score computation involves the dot product $QK^\top$ computed across $H$ attention heads, resulting in a cost of $BHn^{2}d_{k}$. The subsequent weighted value computation, given by $\text{SoftMax}\,(QK^\top)V$, entails multiplying a length-$n$ vector by an $n \times d_{k}$ matrix across $H$ heads, also costing $BHn^{2}d_{k}$. Minor operations---such as scaling by $1/\sqrt{d_k}$, summation operations, and normalization---can be ignored. The attention mechanism computes scores for all cached keys (of length $n$). In Llama 2, a two-layer MLP is employed, featuring a hidden layer of size $d_{ff}$ and utilizing SwiGLU~\cite{SwiGLU} activation during the up-projection.

\begin{equation}
\label{eq:flops_attention}
\text{FLOPs}_{\text{base\_attention}}
= 4Bnd_{\text{model}}^{2} + 2BHn^{2}d_{k}
\end{equation}

\begin{equation}
\label{eq:flops_ffn}
\text{FLOPs}_{\text{FFN}}
= 3Bnd_{\text{model}}d_{ff}
\end{equation}

\textbf{KV-Efficient Attention Mechanism:} We propose a KV-Efficient attention mechanism that optimally manages memory and computation in long-context transformers. The recent window of length $W$ is always preserved in its raw, uncompressed form and is never subject to any transformation. For tokens outside this window, we partition the sequence into non-overlapping chunks of size $C$. Each completed chunk is aggregated into a single compact key--value pair by a 2-layer MLP, denoted as $(\bar{K}, \bar{V})$, which is then processed by a lightweight LSTM gating module. It emits a retention score that exceeds a predefined threshold $\alpha$, to decide whether to keep or forget.

At each inference step, attention is computed over the union of the uncompressed recent window of length $W$, and all retained, compacted chunk tokens $M$. Consequently, the effective memory length for attention becomes $n' = W + M \ll n$, where $n$ is the original sequence length. Notably, whenever the KV cache grows to a certain size, we trigger an aggregation-and-gating step that compresses the stored keys and values.

The total FLOPs~\cite{kaplan2020scaling} per attention layer can be estimated as:
\begin{equation}
\operatorname{FLOPs}_{\text{KV\_Eff}} 
= 4 B n' d_{\text{model}}^{2}
+ 2 B H {n'}^{2} d_{k}
+ \frac{n}{C} B H \left(4 d_{\text{sum}} d_{\text{model}}
+ 4 d_{g} d_{\text{model}}
+ 4 d_{g}^{2}\right).
\end{equation}

Here, $d_{\text{sum}}$ is the MLP-based aggregator hidden dimension. 
The primary computational speedup is achieved through attention, with the throughput (tokens/sec) scaling as
\begin{equation}
\text{Speed up}
= \frac{\operatorname{FLOPs}_{\text{base}} + \operatorname{FLOPs}_{\text{SFF}}}
{\operatorname{FLOPs}_{\text{KV\_Eff}} + \operatorname{FLOPs}_{\text{SFF}}}.
\end{equation}

Replacing or forgetting older spans in this manner significantly reduces both KV memory and bandwidth requirements. The ratio of memory speedup is given by $\frac{n}{W + M}$, reflecting the reduced memory footprint due to compactification and selective retention of historical context.

\section{Experiments}

We evaluate the proposed KV-Efficient framework through both theoretical analysis and empirical experiments. Section~4.1 presents a cost-model analysis between the baseline and our proposed mechanism, quantifying reductions in FLOPs cost, and memory savings. Section~4.2 details our experimental setup and measurement results.

\subsection{Theoretical Results}

We evaluate our method with Hugging Face’s LLaMA-2-7B~\cite{LLAMA2} (with hidden dimension $d_{model} = 4096$, number of layers $L = 32$, attention heads $H = 32$, each with head dimension $d_{k} = 128$, batch size $B = 1$, $d_{ff} = 11008$). The model uses Grouped Query Attention~\cite{ainslie2023gqa} with 8 key--value (KV) heads and operates in bfloat16 precision. Each input sequence has an average length of $n \approx 20{,}000$ tokens, includes vision features, task instruction, robot state, and RGB images resized to $224 \times 224$, across two historical observations and one predicted action. The per-layer, per-token FLOPs for the baseline model Equation~(\ref{eq:flops_attention}) and (\ref{eq:flops_ffn}) is:

\[
\text{FLOPs}_{base} = 1{,}490{,}288{,}640{,}000 \ \text{FLOPs}
\]
\[
\text{FLOPs}_{SFF} = 1{,}351{,}756{,}800{,}000 \ \text{FLOPs}
\]

With the KV-Efficient module, we retain a fixed recent window of size $W = 4096$ covering the instruction, robot state, and current image. The past context is chunked with chunk size $C = 3136$, MLP aggregator has hidden dimension $d_{\text{sum}} = 128$, and the gate is an LSTM-based retention module with an average retention ratio $r = 0.687$, operating on a hidden size of $d_g = 128$.

Substituting into Equation (6), the KV-Efficient attention cost is reduced to:

\[
\text{FLOPs}_{KV\_Eff} \approx 412{,}871{,}032{,}320 \ \text{FLOPs}
\]

This corresponds to an attention-level speedup computed using Equation~(7):

\[
\text{Speed Up} \approx 1.61 \times
\]

The ratio of memory speedup is $2.44 \times$

\subsection{Experiment:}

\textbf{Dataset.} We employ the Open X-Embodiment dataset~\cite{embodimentcollaboration2025openxembodimentroboticlearning}, which offers over 500,000 demonstrations encompassing 22 robot embodiments and more than 500 tasks. Each sample consists of RGB images, language instructions, robot states, and low-level continuous actions, thus providing rich multimodal input-output mappings. This diversity is essential for generalizing VLA models across various tasks, embodiments, and sensor modalities.

\textbf{Simulation Environment.} The RLBench environment~\cite{james2019rlbenchrobotlearningbenchmark}, built on the CoppeliaSim simulator~\cite{CoppeliaSim}, serves as the primary simulation platform. RLBench features over 100 diverse tasks, with approximately 1,000 expert demonstrations per task, yielding more than 100,000 trajectories. Each trajectory includes multimodal observations such as multi-view RGB images, depth maps, segmentation masks, and proprioceptive states.

\textbf{Experimental Settings.} We fine-tuned the proposed KV-efficient memory mechanism exclusively on the HybridVLA model. Both KV-cache chunking and LSTM-based gate are also transferable to other VLA architectures, including OpenVLA~\cite{kim2024openvla} and CogACT~\cite{li2024cogact}. The model was initialized with the pretrained base VLM, prism-dinosiglip-224px+7b~\cite{karamcheti2024prismatic}. The vision-language action model consists of a DINO~\cite{oquab2023dinov2} SigLIP~\cite{zhai2023sigmoid} ViT-Large vision backbone, pretrained with ViT-SO400M-14-SigLIP~\cite{big_vision} weights and implemented using the Hugging Face Transformers library. For the LLM backbone, LLaMA-2 7B (“llama2-7b-pure”) is employed. The model outputs are produced via a fused GELU ~\cite{hendrycks2016gaussian} multi-layer perceptron (MLP) head, without the use of an additional alignment module.

Owing to computational resource constraints, we perform illustrative fine-tuning using mixed precision training on two NVIDIA H800 GPUs. Figure 3 presents the loss curve for the HybridVLA baseline augmented with our KV-efficient module. Notably, the loss exhibits a decrease within the first 100 iterations, followed by stable convergence. This indicates that our proposed mechanism remains compatible with HybridVLA training dynamics.

\begin{figure}[t]
    \centering
    \includegraphics[width=0.65\linewidth]{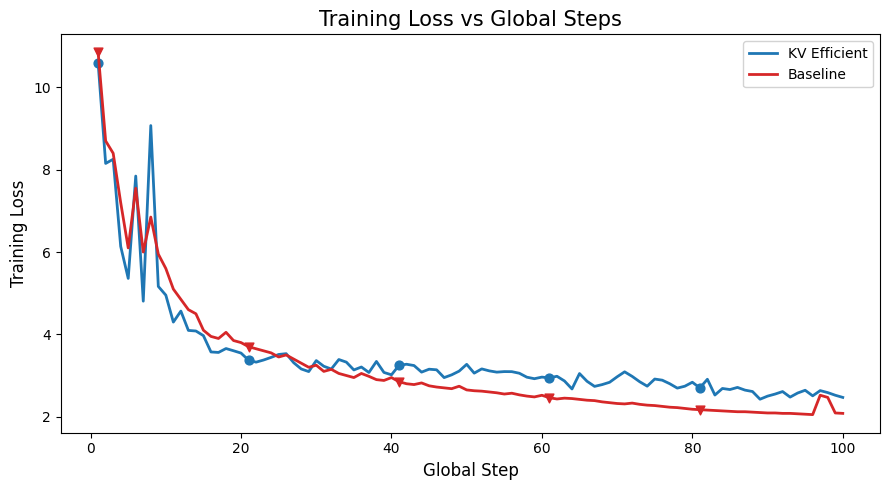}
    \caption{Training Loss for KV-Efficient module}
\end{figure}

To isolate and quantify computational efficiency in comparison to our theoretical analysis, we evaluate the computational cost under long-context conditions using synthetic token sequences. We measure the per-token computational cost incurred during inference, where the model auto-regressively generates action vectors. For inference speed evaluation, measurements are conducted on a single robotic task to ensure consistency across experiments.

\begin{table}[h!]
\centering
\caption{Comparison of KV-Efficient. All methods in the multi-task setting}
\begin{tabular}{lccc}
\hline
Models & Speed up $\times$ & Infer. Speed (Hz) & Total-FLOPs (T) \\
\hline
OpenVLA (7B)~\cite{kim2024openvla} & 1.00 & 6.3 & 2.37 \\
OpenVLA-KV-Efficient & 1.22 & 7.6 & 1.94 \\
CogACT (7B)~\cite{li2024cogact} & 1.00 & 9.8 & 2.41 \\
CogACT-KV-Efficient & 1.33 & 13.8 & 1.81 \\
HybridVLA (7B)~\cite{liu2025hybridvla} & 1.00 & 5.8 & 2.73 \\
HybridVLA-KV-Efficient & 1.47 & 8.3 & 1.85 \\
\hline
\end{tabular}
\end{table}

As shown in Table~1, our method was evaluated on three baseline VLA models and their KV-efficient variants.

\textbf{Inference Speed and Memory.} The KV-Efficient approach reduces average total FLOPs by approximately 24.6\% compared to the baselines. Importantly, all variants maintain the same parameter count, confirming that efficiency gains arise from computational savings rather than reduced model size.

The inference throughput is significantly improved with KV-efficient caching: OpenVLA achieves an inference speed of 7.6~Hz, CogACT with 13.8~Hz, and HybridVLA with 8.3~Hz. Though these speeds fall slightly below the theoretical maximum due to memory compression overhead, they represent substantial speedups over the original models. On average, our model can increase the inference speed by about 1.34$\times$. KV-efficient variants also reduce memory usage by approximately 1.87$\times$ in KV cache storage, supporting faster inference with lower resource demands.

These results clearly demonstrate the effectiveness of our model in improving inference speed and memory efficiency. As the input sequence length increases, the benefits become even more pronounced, particularly in the attention computation. However, due to compilation constraints, accuracy and inference speed were evaluated on only a single benchmark task, which may not fully reflect the general performance of the model.

\section{Summary}

In this paper, we present the KV-Efficient module, a lightweight memory mechanism designed to improve the inference efficiency and scalability of Vision--Language--Action models by chunking and compressing historical key--value caches. Our approach employs an LSTM-based gating module to selectively retain relevant chunks while minimizing redundant memory usage. We implement and evaluate the KV-Efficient mechanism within recently established VLA architectures, leveraging large-scale and diverse datasets from Open X-Embodiment. Through both theoretical cost-model analysis and empirical benchmarking, we observe reductions in attention FLOPs and KV cache memory usage, resulting in inference speedups.

Looking ahead, we plan to conduct detailed ablation studies on chunk size and retention gating strategies, and to deploy our approach in real-world, closed-loop robotic settings to assess both responsiveness and safety.

\bibliographystyle{plain}
\bibliography{references}

\end{document}